\documentclass[sigconf]{acmart}
\usepackage{amsmath}
\usepackage{bm}
\usepackage{commath}
\usepackage{multirow}
\usepackage{booktabs}
\AtBeginDocument{%
  \providecommand\BibTeX{{%
    \normalfont B\kern-0.5em{\scshape i\kern-0.25em b}\kern-0.8em\TeX}}}

\copyrightyear{2021} 
\acmYear{2021} 
\setcopyright{acmcopyright}\acmConference[MMPT '21]{Proceedings of the 2021 Workshop on Multi-Modal Pre-Training for Multimedia Understanding}{August 21, 2021}{Taipei, Taiwan}
\acmBooktitle{Proceedings of the 2021 Workshop on Multi-Modal Pre-Training for Multimedia Understanding (MMPT '21), August 21, 2021, Taipei, Taiwan}
\acmPrice{15.00}
\acmDOI{10.1145/3463945.3469056}
\acmISBN{978-1-4503-8530-5/21/08}

\begin{document}

\title{Residual Recurrent CRNN for End-to-End Optical Music Recognition on Monophonic Scores}


\author{Aozhi Liu}
\email{liuaozhi201310@163.com}
\affiliation{%
  \institution{Ping An Technology(Shenzhen) Co.,Ltd}
  \country{China}
}

\author{Lipei Zhang}
\authornote{Corresponding Author.}
\email{oscarzlp@outlook.com}
\affiliation{%
  \institution{University College London}
  \country{United Kingdom}
}
\author{Yaqi Mei}
\email{meiyaqi381@pingan.com.cn}
\affiliation{%
  \institution{Ping An Technology(Shenzhen) Co.,Ltd}
  \country{China}
}

\author{Baoqiang Han}
\email{hanbaoqiang038@pingan.com.cn}
\affiliation{%
  \institution{Ping An Technology(Shenzhen) Co.,Ltd}
  \country{China}
}

\author{Zifeng Cai}
\email{caizifeng709@pingan.com.cn}
\affiliation{%
  \institution{Ping An Technology(Shenzhen) Co.,Ltd}
  \country{China}
}

\author{Zhaohua Zhu}
\email{zhuzhaohua582@pingan.com.cn}
\affiliation{%
  \institution{Ping An Technology(Shenzhen) Co.,Ltd}
  \country{China}
}

\author{Jing Xiao}
\email{xiaojing661@pingan.com.cn}
\affiliation{%
  \institution{Ping An Technology(Shenzhen) Co.,Ltd	}
  \country{China}
}
\begin{abstract}
One of the challenges of the Optical Music Recognition task is to transcript the symbols of the camera-captured images into digital music notations. Previous end-to-end model which was developed as a Convolutional Recurrent Neural Network does not explore sufficient contextual information from full scales and there is still a large room for improvement. We propose an innovative  framework that combines a block of Residual Recurrent Convolutional Neural Network with a recurrent Encoder-Decoder network to map a sequence of monophonic music symbols corresponding to the notations present in the image. The Residual Recurrent Convolutional block can improve the ability of the model to enrich the context information. The experiment results are benchmarked against a publicly available dataset called CAMERA-PRIMUS, which demonstrates that our approach surpass the state-of-the-art end-to-end method using Convolutional Recurrent Neural Network.
\end{abstract}






\maketitle

\section{Introduction}\label{sec:introduction}
Optical Music Recognition(OMR) is an area of research that investigates how to decode music notations from images. Most of the early studies of Optical Music Recognition have focused on a multi-stage fashion\cite{rebelo2012optical}. The complete workflow of the methods normally includes an image prepossessing and image binarization initially\cite{calvo2017pixel}\cite{vo2016mrf}. Following this step is typically a staff-line detection and removal process. It is worth mentioning that the staff-line detection and removal part is critically important since the failure of this step will result in the failure of the rest parts\cite{fornes2011icdar}\cite{su2012effective}. In the next step, the symbol classification is typically processed to distinguish the musical meaning of each symbol\cite{lee2016handwritten}\cite{calvo2017recognition}. Once these symbols are classified into different categories, they have to be put together to build up the semantic meaning of the music scores\cite{raphael2011new}. In recent years, as an early attempt of end-to-end system in Optical Music Recognition, Jorge Calvo-Zaragoze et al. implemented a Convolutional Recurrent Neural Network(CRNN) framework proposed by Shi et al.\cite{shi2016end} for scene text recognition to solve the OMR problem with deep learning methods\cite{calvo2018end}. The basic CRNN structure includes several convolutional blocks for extracting features of the images followed by a recurrent block for dealing with the sequential task regarding to the features as well as the output symbols. 

However, the feed-forward only convolutional block for the whole CRNN architecture has limitation on the performance of extraction and integration on the contextual information. Inspired by the deep residual model\cite{he2016deep} and the RCNN structure applying in several computer vision tasks\cite{pinheiro2014recurrent}\cite{liang2015recurrent}, we have proposed a Recurrent Residual Convolutional Neural Network for our end-to-end Optical Music Recognition task. The recurrent neural network has been implemented as the subsequent block of the recurrent residual convolutional block for the entire end-to-end system. The recurrent residual convolutional block ensures better feature representation and wider range of contextual information derivation for the OMR system. Moreover, we have evaluated the performance of our proposed model with the dataset called CAMERA-PRIMUS\cite{calvo2018camera} containing both the printed image and as well as the image with distortion. The experiments demonstrate that the results of our end-to-end framework with recurrent residual convolutional blocks outperform the results of the vanilla CRNN framework\cite{calvo2018end}\cite{calvo2018camera}, which indicates that our proposed model have achieved the state-of-the-art performance in end-to-end monophonic score recognition system.

\section{Approach}\label{sec:approach}
\subsection{Data Representation}
	
It is well known that the tasks of deep learning require dataset with sufficient size as well as good quality. For these requirements, we implement the \textit{Camera-based Printed Images of Music Staves}(Camera-PrIMuS)\cite{calvo2018camera} dataset for our end-to-end training system. This dataset contains 87678 clean images of music score and the same number of synthetic distorted images to resemble the camera-based scenario.
	
\begin{figure*}[htb]
	\centering
	\includegraphics[scale=0.48]{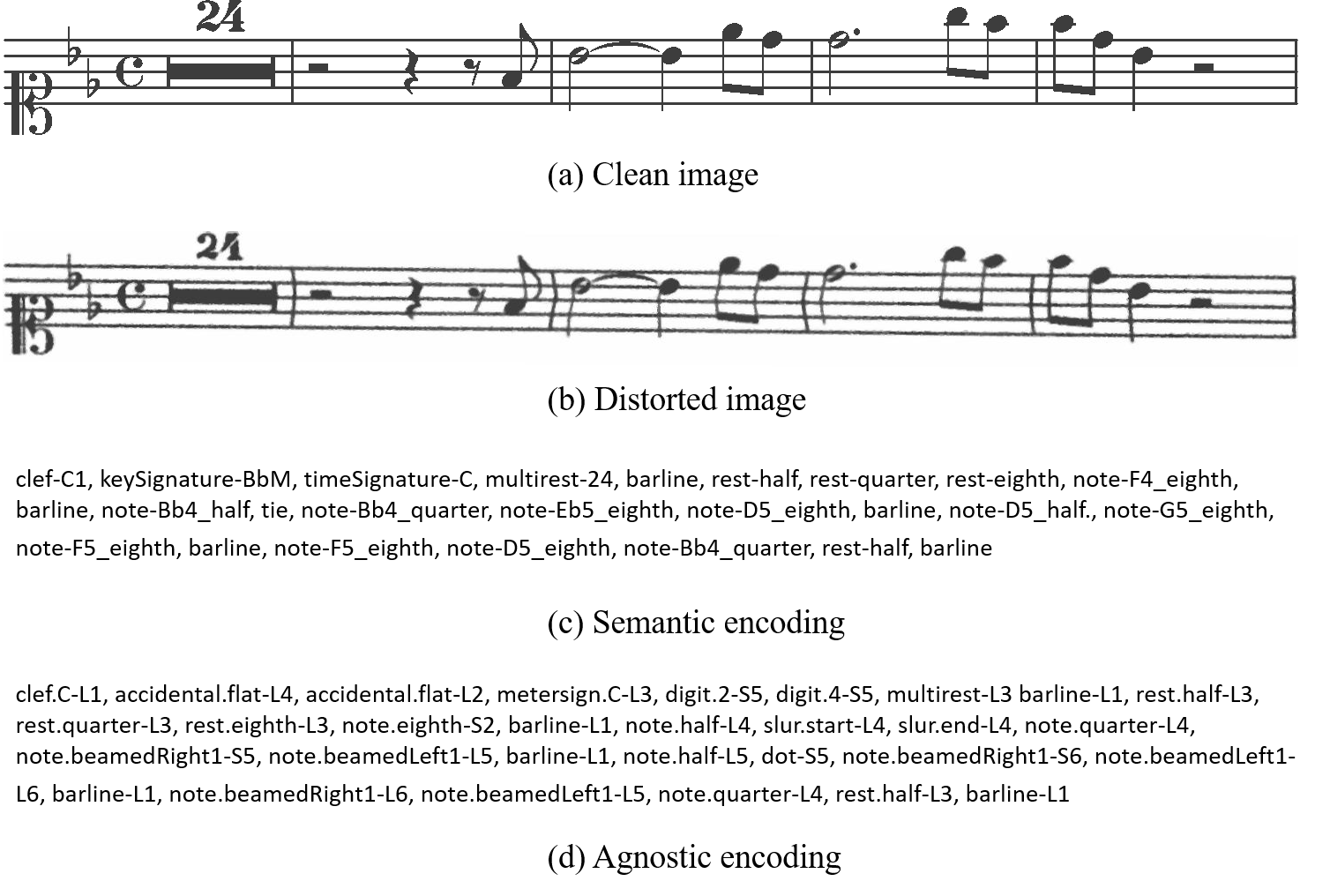}
	\caption{Camera-PrIMus incipit contents example. (a)Example of the clean image. (b)Example of the distorted image (c)Example of semantic encoding (d)Example of agnostic encoding}
	\label{fig:figure1}
\end{figure*}

For each image, there are two kinds of representations to fulfill the end-to-end training purpose of the deep neural network. The examples are shown in  Fig. \ref{fig:figure1}. One of the representations is called semantic encoding which includes symbols with musical meaning. For instance, a \textsuperscript{b}B major key signature (see Fig. \ref{fig:figure1}(c)) is notated with the musical meaning "keySignature". On the other hand, the agnostic encoding (see Fig.  \ref{fig:figure1}(d)) contains a list of graphic symbols without musical meaning, e.g. a A Major key signature is represented as a sequence of three \textit{sharp} symbols. 

\begin{figure*}[htb]
	\centering
	\includegraphics[scale=0.5]{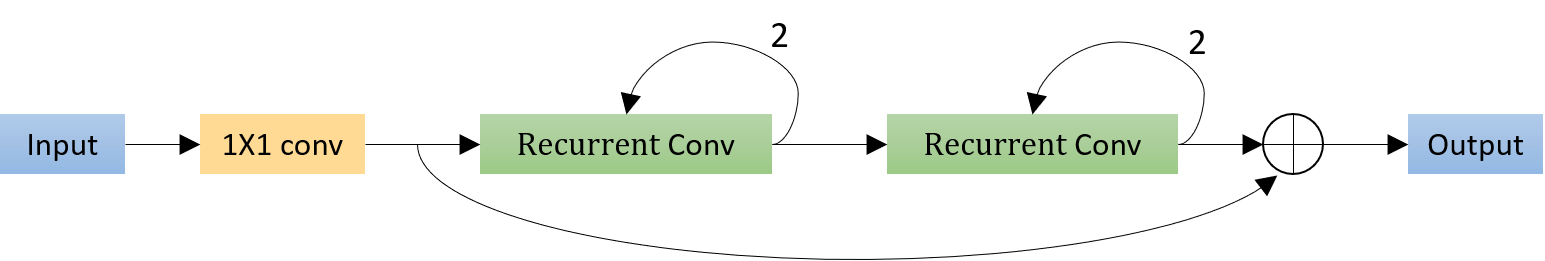}
	\caption{The Structure of Residual Recurrent Convolutional Block.}
	\label{fig:figure2}
\end{figure*}
	
\begin{figure*}[htb]
	\centering
	\includegraphics[scale=0.5]{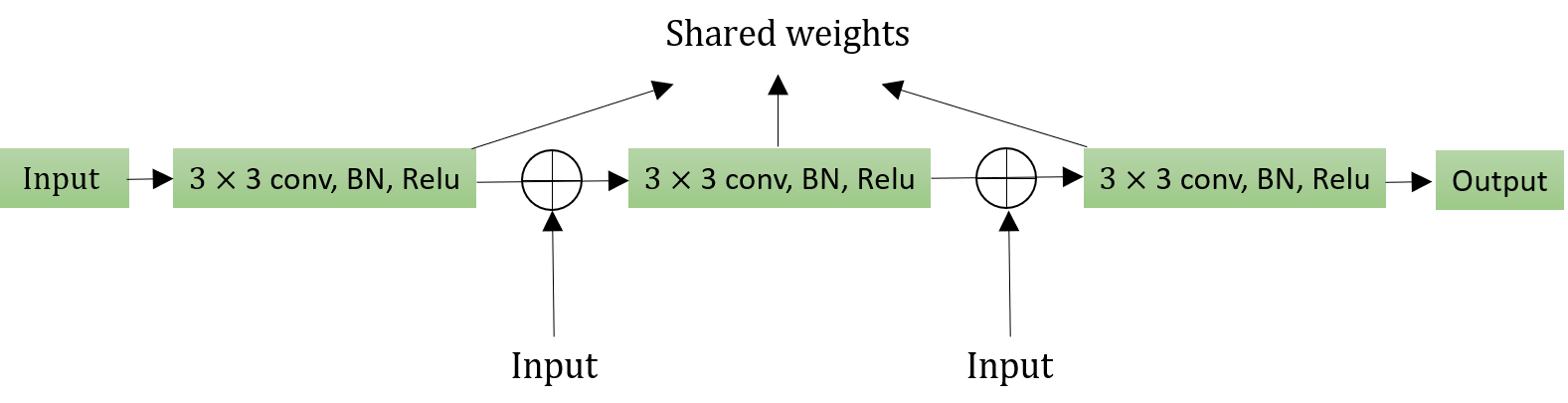}
	\caption{Unfolded Recurrent Convolutional Unit.}
	\label{fig:figure3}
\end{figure*}

\begin{figure*}[htb]
	\centering
	\includegraphics[scale=0.5]{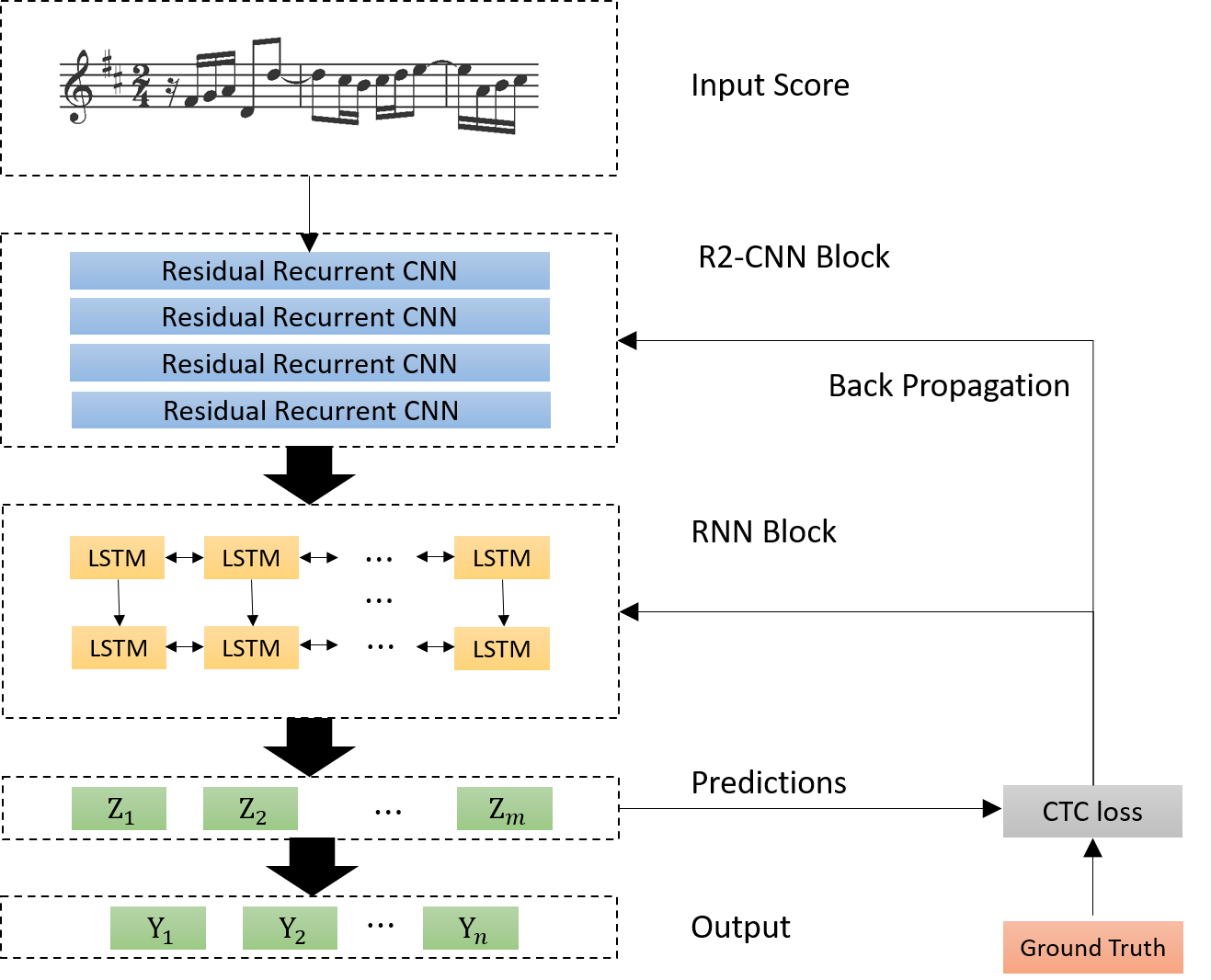}
	\caption{Graphical scheme of the R2-CRNN end-to-end model.}
	\label{fig:figure4}
\end{figure*}

\begin{table*}[!h]
	\centering
	\caption{Average SER (\%) / ER (\%) values in all possible combinations of training and evaluation conditions.}
	\begin{tabular}{c|c|c|cccc}
		\toprule[1pt]
		& &  & \multicolumn{4}{c}{\textbf{Evaluation}} \\
		\midrule[0.5pt]
		& &  & \multicolumn{2}{c}{\textbf{Clean}} & \multicolumn{2}{c}{\textbf{Distortions}}\\
		\midrule[0.5pt]
		& & & Agnostic & Semantic & Agnostic & Semantic \\
		\midrule[0.5pt]
		\multirow{4}{*}{Training} & \multirow{2}{*}{Clean} & CRNN results & 1.10/21.7 & 0.80/12.5 & 44.3/95.1 & 59.7/97.9\\
		&  & \textbf{Our results} & \textbf{0.59/16.2} & \textbf{0.56/13.4} & \textbf{7.63/20.9} & \textbf{38.19/91.0}\\
		& \multirow{2}{*}{Distortions} & CRNN results & 1.40/24.9 & 3.3/44.6 & 1.6/24.7 & 3.4/38.3\\
		&  & \textbf{Our results} & \textbf{0.68/18.7} & \textbf{1.06/25.4} & \textbf{1.01/27.8} & \textbf{1.50/35.8}\\
		\bottomrule[1pt]
	\end{tabular}
	\label{tab:table2}
\end{table*}

\subsection{Model Architecture}
 In this section, we describe our end-to-end framework containing the feature extraction component integrating the deep residual model\cite{he2016deep} and the Recurrent Convolutional Neural Networks (RCNN) model. The complete framework is named as Residual Recurrent Convolutional Recurrent Neural Networks (R2-CRNN).	

\subsubsection{Residual Recurrent Block}
	
The residual recurrent block is an idea integrating the residual model\cite{he2016deep} as well as the recurrent convolutional  neural network (RCNN)\cite{liang2015recurrent} to improve the accuracy of the monophonic optical music recognition task. The residual recurrent convolutional block is shown in Fig. \ref{fig:figure2}. The $1\times 1$ convolutional operation\cite{lin2013network} is applied to reduce the number of channels of the input. Followed by this operation is 2 recurrent units using convolution. The recurrent convolutional unites are able to enrich the extracted output features by feature accumulation with the recurrent mechanism. The detailed structure of the  recurrent convolutional units is shown in Fig. \ref{fig:figure3}. The channels reduction results of the image input will be taken as the input of the first recurrent convolutional unit. In the first time step, a kernel with the size of $3\times 3$ is used for convolution, followed by a Batch Normalization\cite{ioffe2015batch} process to increase the training efficiency and Rectified Linear Unit\cite{glorot2011deep} for activation. Then, the addition of the input of the recurrent convolutional unit and the sequential operation combined with $3\times 3$ convolution, Batch Normalization and Rectified Linear Unit activation will be processed twice to derive the output of the recurrent convolutional unit. As the final step of the residual recurrent convolutional block, a residual mechanism is applied so that the output of the block is obtained.

\subsubsection{End-to-end Architecture}
	
Some previous work has demonstrated that end-to-end recognition of monophonic music score is plausible\cite{calvo2017end}\cite{calvo2017recognition}\cite{calvo2018camera}. The end-to-end problem is constructed by an input staff image as well as its corresponding sequence of music symbols.
    
The graphic scheme of the end-to-end framework is shown in Fig. \ref{fig:figure4}. An image of monophonic staff is taken as the input of the Residual Recurrent Convolutional Recurrent Neural Network (R2-CRNN) model without any preprocessing steps such as symbol segmentation or staff-line removal. The R2-CRNN model consists of two blocks: the Residual Recurrent Convolutional Neural Network Block (R2-CNN) and the Recurrent Neural Network Block (RNN). Firstly, the image is processed by the R2-CNN block for feature extraction. Secondly, the features extracted by the R2-CNN block are fed into the RNN block, which finally converts the image into a sequence of music symbols. Moreover, a CTC loss function\cite{graves2006connectionist} is applied so that there's no need for framewise labelling of the input data. However, the CTC loss is applied only at the training stage. In the prediction step, the framewise R2-CRNN output can be decoded into a sequence of music symbols.
    

\subsubsection{Implementation details}
 Firstly, the input image is rescaled at a fixed height of 128 pixels. Then, this input will be processed through a R2-CNN block consisting of four residual recurrent units. Each residual recurrent unit has a $1\times 1$ convolution operator, two recurrent convolution operators with $3\times 3$ kernel size and a maxpooling down-sampling operator with $2\times 2$ window size. The output of the R2-CNN block is then taken as the input of two bi-directional LSTM\cite{hochreiter1997long} layers of 256 neurons. Considering both the input features and the modeling of the musical representation, the bi-directioanl LSTM will produce the discrete musical symbol sequence. Finally, a fully-connected layer with a softmax activation function is used for the classification of the music symbol class of each frame. To learn the model, the parameters of the R2-CNN block as well as the Recurrent block are updated repeatedly through the back-propagation using Adam optimizer\cite{kingma2014adam} to minimize the CTC loss function. The batch size to train the model is set to be 16. 
	
At the stage of prediction, the decoding process can be performed greedily due to the implementation of CTC loss function. If a predicted symbol of a certain frame is the same as the previous one, it is considered the symbol that covers more than one frame. Hence, this symbol will not be appended to the decoding sequence of symbols. On the other hand, if the predicted symbol is different from the previous one or the predicted symbol is the blank symbol, it will be concatenated to the music symbol decoding sequence.  

\section{Experiment}
\label{sec:typestyle}

\subsection{Experiment step}
	
We would like to perform experiments to demonstrate that our model has achieved the state-of-the-art performances in the field of monophonic optical music recognition. For this purpose, the evaluation approach that has been implemented in the previous end-to-end OMR tasks\cite{calvo2018end}\cite{calvo2018camera} is implemented in our work. In this way, we use two quantitative metrics that take into account both the sequence aspect and the symbol aspect:
	
\begin{itemize}
	\item Sequence Error Rate (ER) (\%): ratio of incorrectly predicted sequences (at least one error)/
	\item Symbol Error Rate (SER) (\%): the average number of editing operations such as insertions, deletions, or substitutions that are necessary to generate the reference sequence from the one predicted by the model, normalized by its length.
\end{itemize}
	
In Calvo-Zaragoza J.'s paper\cite{calvo2018camera}, for both the clean images and the distorted images the ER as well as the SER values are evaluated on the results with both the agnostic labelling and the semantic notation. For comparing our results to this work, which is the most up-to-date end-to-end framework in OMR previous to our study, we evaluate all metrics in this paper to show the state-of-the-art performance of our model.
	
\subsection{Performance}

In this section, we show the results of our experiments that have compared with the results of Calvo-Zaragoza J.'s work\cite{calvo2018camera}. We consider three different data partitions of the dataset that 80\% of the data for training while 10\% for validation. The remaining 10\% is used for test. 
	
Calvo-Zaragoza J.'s work\cite{calvo2018camera} reported all possible results regarding the training and evaluation scenarios. To comprehensively evaluate the performances of our model, we also perform the test on all of these conditions. The detailed comparison is shown in the Table. \ref{tab:table2}. The most notable difference is in the condition that training on clean images while evaluation on distorted images with agnostic notation. The SER (\%) value and the ER (\%) value are 44.3 and 95.1 respectively when using CRNN model. With our model, the SER (\%) value and the ER (\%) value are only 7.63 and 20.9. Actually, in all conditions the results of our model surpass the performances of the Convolutional Recurrent Neural Network (CRNN) model implemented by Calvo-Zaragoza J. which reflect that our R2-CRNN model has significant improvement of the up-to-date CRNN model on the monophonic Optical Music Recognition tasks.

\section{Conclusion}
In this work, we have proposed an innovative end-to-end framework for Optical Music Recognition consisting a Residual Recurrent Convolution block as well as a Recurrent Neural Network block. Compared to early end-to-end model, our improvement enhanced the capability of the kernel filters to capture statistical features in the context of the symbols.The results of experiments demonstrate that our model outperform the up-to-date model(CRNN) architecture, which have achieved the state-of-the-art results of the tasks of monophonic Optical Music Recognition.


\bibliographystyle{ACM-Reference-Format}
\bibliography{sample-base}


\begin{thebibliography}{21}


\ifx \showCODEN    \undefined \def \showCODEN     #1{\unskip}     \fi
\ifx \showDOI      \undefined \def \showDOI       #1{#1}\fi
\ifx \showISBNx    \undefined \def \showISBNx     #1{\unskip}     \fi
\ifx \showISBNxiii \undefined \def \showISBNxiii  #1{\unskip}     \fi
\ifx \showISSN     \undefined \def \showISSN      #1{\unskip}     \fi
\ifx \showLCCN     \undefined \def \showLCCN      #1{\unskip}     \fi
\ifx \shownote     \undefined \def \shownote      #1{#1}          \fi
\ifx \showarticletitle \undefined \def \showarticletitle #1{#1}   \fi
\ifx \showURL      \undefined \def \showURL       {\relax}        \fi
\providecommand\bibfield[2]{#2}
\providecommand\bibinfo[2]{#2}
\providecommand\natexlab[1]{#1}
\providecommand\showeprint[2][]{arXiv:#2}

\bibitem[\protect\citeauthoryear{Calvo-Zaragoza, Gallego, and
  Pertusa}{Calvo-Zaragoza et~al\mbox{.}}{2017a}]%
        {calvo2017recognition}
\bibfield{author}{\bibinfo{person}{Jorge Calvo-Zaragoza},
  \bibinfo{person}{Antonio-Javier Gallego}, {and} \bibinfo{person}{Antonio
  Pertusa}.} \bibinfo{year}{2017}\natexlab{a}.
\newblock \showarticletitle{Recognition of handwritten music symbols with
  convolutional neural codes}. In \bibinfo{booktitle}{\emph{2017 14th IAPR
  International Conference on Document Analysis and Recognition (ICDAR)}},
  Vol.~\bibinfo{volume}{1}. IEEE, \bibinfo{pages}{691--696}.
\newblock


\bibitem[\protect\citeauthoryear{Calvo-Zaragoza and Rizo}{Calvo-Zaragoza and
  Rizo}{2018a}]%
        {calvo2018camera}
\bibfield{author}{\bibinfo{person}{Jorge Calvo-Zaragoza} {and}
  \bibinfo{person}{David Rizo}.} \bibinfo{year}{2018}\natexlab{a}.
\newblock \showarticletitle{Camera-PrIMuS: Neural End-to-End Optical Music
  Recognition on Realistic Monophonic Scores.}. In
  \bibinfo{booktitle}{\emph{ISMIR}}. \bibinfo{pages}{248--255}.
\newblock


\bibitem[\protect\citeauthoryear{Calvo-Zaragoza and Rizo}{Calvo-Zaragoza and
  Rizo}{2018b}]%
        {calvo2018end}
\bibfield{author}{\bibinfo{person}{Jorge Calvo-Zaragoza} {and}
  \bibinfo{person}{David Rizo}.} \bibinfo{year}{2018}\natexlab{b}.
\newblock \showarticletitle{End-to-end neural optical music recognition of
  monophonic scores}.
\newblock \bibinfo{journal}{\emph{Applied Sciences}} \bibinfo{volume}{8},
  \bibinfo{number}{4} (\bibinfo{year}{2018}), \bibinfo{pages}{606}.
\newblock


\bibitem[\protect\citeauthoryear{Calvo-Zaragoza, Valero-Mas, and
  Pertusa}{Calvo-Zaragoza et~al\mbox{.}}{2017b}]%
        {calvo2017end}
\bibfield{author}{\bibinfo{person}{Jorge Calvo-Zaragoza},
  \bibinfo{person}{Jose~J Valero-Mas}, {and} \bibinfo{person}{Antonio
  Pertusa}.} \bibinfo{year}{2017}\natexlab{b}.
\newblock \showarticletitle{End-to-end optical music recognition using neural
  networks}. In \bibinfo{booktitle}{\emph{Proceedings of the 18th International
  Society for Music Information Retrieval Conference, ISMIR}}.
  \bibinfo{pages}{23--27}.
\newblock


\bibitem[\protect\citeauthoryear{Calvo-Zaragoza, Vigliensoni, and
  Fujinaga}{Calvo-Zaragoza et~al\mbox{.}}{2017c}]%
        {calvo2017pixel}
\bibfield{author}{\bibinfo{person}{Jorge Calvo-Zaragoza},
  \bibinfo{person}{Gabriel Vigliensoni}, {and} \bibinfo{person}{Ichiro
  Fujinaga}.} \bibinfo{year}{2017}\natexlab{c}.
\newblock \showarticletitle{Pixel-wise binarization of musical documents with
  convolutional neural networks}. In \bibinfo{booktitle}{\emph{2017 Fifteenth
  IAPR International Conference on Machine Vision Applications (MVA)}}. IEEE,
  \bibinfo{pages}{362--365}.
\newblock


\bibitem[\protect\citeauthoryear{Fornes, Dutta, Gordo, and Llados}{Fornes
  et~al\mbox{.}}{2011}]%
        {fornes2011icdar}
\bibfield{author}{\bibinfo{person}{Alicia Fornes}, \bibinfo{person}{Anjan
  Dutta}, \bibinfo{person}{Albert Gordo}, {and} \bibinfo{person}{Josep
  Llados}.} \bibinfo{year}{2011}\natexlab{}.
\newblock \showarticletitle{The ICDAR 2011 music scores competition: Staff
  removal and writer identification}. In \bibinfo{booktitle}{\emph{2011
  International Conference on Document Analysis and Recognition}}. IEEE,
  \bibinfo{pages}{1511--1515}.
\newblock


\bibitem[\protect\citeauthoryear{Glorot, Bordes, and Bengio}{Glorot
  et~al\mbox{.}}{2011}]%
        {glorot2011deep}
\bibfield{author}{\bibinfo{person}{Xavier Glorot}, \bibinfo{person}{Antoine
  Bordes}, {and} \bibinfo{person}{Yoshua Bengio}.}
  \bibinfo{year}{2011}\natexlab{}.
\newblock \showarticletitle{Deep sparse rectifier neural networks}. In
  \bibinfo{booktitle}{\emph{Proceedings of the fourteenth international
  conference on artificial intelligence and statistics}}.
  \bibinfo{pages}{315--323}.
\newblock


\bibitem[\protect\citeauthoryear{Graves, Fern{\'a}ndez, Gomez, and
  Schmidhuber}{Graves et~al\mbox{.}}{2006}]%
        {graves2006connectionist}
\bibfield{author}{\bibinfo{person}{Alex Graves}, \bibinfo{person}{Santiago
  Fern{\'a}ndez}, \bibinfo{person}{Faustino Gomez}, {and}
  \bibinfo{person}{J{\"u}rgen Schmidhuber}.} \bibinfo{year}{2006}\natexlab{}.
\newblock \showarticletitle{Connectionist temporal classification: labelling
  unsegmented sequence data with recurrent neural networks}. In
  \bibinfo{booktitle}{\emph{Proceedings of the 23rd international conference on
  Machine learning}}. \bibinfo{pages}{369--376}.
\newblock


\bibitem[\protect\citeauthoryear{He, Zhang, Ren, and Sun}{He
  et~al\mbox{.}}{2016}]%
        {he2016deep}
\bibfield{author}{\bibinfo{person}{Kaiming He}, \bibinfo{person}{Xiangyu
  Zhang}, \bibinfo{person}{Shaoqing Ren}, {and} \bibinfo{person}{Jian Sun}.}
  \bibinfo{year}{2016}\natexlab{}.
\newblock \showarticletitle{Deep residual learning for image recognition}. In
  \bibinfo{booktitle}{\emph{Proceedings of the IEEE conference on computer
  vision and pattern recognition}}. \bibinfo{pages}{770--778}.
\newblock


\bibitem[\protect\citeauthoryear{Hochreiter and Schmidhuber}{Hochreiter and
  Schmidhuber}{1997}]%
        {hochreiter1997long}
\bibfield{author}{\bibinfo{person}{Sepp Hochreiter} {and}
  \bibinfo{person}{J{\"u}rgen Schmidhuber}.} \bibinfo{year}{1997}\natexlab{}.
\newblock \showarticletitle{Long short-term memory}.
\newblock \bibinfo{journal}{\emph{Neural computation}} \bibinfo{volume}{9},
  \bibinfo{number}{8} (\bibinfo{year}{1997}), \bibinfo{pages}{1735--1780}.
\newblock


\bibitem[\protect\citeauthoryear{Ioffe and Szegedy}{Ioffe and Szegedy}{2015}]%
        {ioffe2015batch}
\bibfield{author}{\bibinfo{person}{Sergey Ioffe} {and}
  \bibinfo{person}{Christian Szegedy}.} \bibinfo{year}{2015}\natexlab{}.
\newblock \showarticletitle{Batch normalization: Accelerating deep network
  training by reducing internal covariate shift}.
\newblock \bibinfo{journal}{\emph{arXiv preprint arXiv:1502.03167}}
  (\bibinfo{year}{2015}).
\newblock


\bibitem[\protect\citeauthoryear{Kingma and Ba}{Kingma and Ba}{2014}]%
        {kingma2014adam}
\bibfield{author}{\bibinfo{person}{Diederik~P Kingma} {and}
  \bibinfo{person}{Jimmy Ba}.} \bibinfo{year}{2014}\natexlab{}.
\newblock \showarticletitle{Adam: A method for stochastic optimization}.
\newblock \bibinfo{journal}{\emph{arXiv preprint arXiv:1412.6980}}
  (\bibinfo{year}{2014}).
\newblock


\bibitem[\protect\citeauthoryear{Lee, Son, Oh, and Kwak}{Lee
  et~al\mbox{.}}{2016}]%
        {lee2016handwritten}
\bibfield{author}{\bibinfo{person}{Sangkuk Lee}, \bibinfo{person}{Sung~Joon
  Son}, \bibinfo{person}{Jiyong Oh}, {and} \bibinfo{person}{Nojun Kwak}.}
  \bibinfo{year}{2016}\natexlab{}.
\newblock \showarticletitle{Handwritten music symbol classification using deep
  convolutional neural networks}. In \bibinfo{booktitle}{\emph{2016
  International Conference on Information Science and Security (ICISS)}}. IEEE,
  \bibinfo{pages}{1--5}.
\newblock


\bibitem[\protect\citeauthoryear{Liang and Hu}{Liang and Hu}{2015}]%
        {liang2015recurrent}
\bibfield{author}{\bibinfo{person}{Ming Liang} {and} \bibinfo{person}{Xiaolin
  Hu}.} \bibinfo{year}{2015}\natexlab{}.
\newblock \showarticletitle{Recurrent convolutional neural network for object
  recognition}. In \bibinfo{booktitle}{\emph{Proceedings of the IEEE conference
  on computer vision and pattern recognition}}. \bibinfo{pages}{3367--3375}.
\newblock


\bibitem[\protect\citeauthoryear{Lin, Chen, and Yan}{Lin et~al\mbox{.}}{2013}]%
        {lin2013network}
\bibfield{author}{\bibinfo{person}{Min Lin}, \bibinfo{person}{Qiang Chen},
  {and} \bibinfo{person}{Shuicheng Yan}.} \bibinfo{year}{2013}\natexlab{}.
\newblock \showarticletitle{Network in network}.
\newblock \bibinfo{journal}{\emph{arXiv preprint arXiv:1312.4400}}
  (\bibinfo{year}{2013}).
\newblock


\bibitem[\protect\citeauthoryear{Pinheiro and Collobert}{Pinheiro and
  Collobert}{2014}]%
        {pinheiro2014recurrent}
\bibfield{author}{\bibinfo{person}{Pedro Pinheiro} {and} \bibinfo{person}{Ronan
  Collobert}.} \bibinfo{year}{2014}\natexlab{}.
\newblock \showarticletitle{Recurrent convolutional neural networks for scene
  labeling}. In \bibinfo{booktitle}{\emph{International conference on machine
  learning}}. \bibinfo{pages}{82--90}.
\newblock


\bibitem[\protect\citeauthoryear{Raphael and Wang}{Raphael and Wang}{2011}]%
        {raphael2011new}
\bibfield{author}{\bibinfo{person}{Christopher Raphael} {and}
  \bibinfo{person}{Jingya Wang}.} \bibinfo{year}{2011}\natexlab{}.
\newblock \showarticletitle{New Approaches to Optical Music Recognition.}. In
  \bibinfo{booktitle}{\emph{ISMIR}}. \bibinfo{pages}{305--310}.
\newblock


\bibitem[\protect\citeauthoryear{Rebelo, Fujinaga, Paszkiewicz, Marcal, Guedes,
  and Cardoso}{Rebelo et~al\mbox{.}}{2012}]%
        {rebelo2012optical}
\bibfield{author}{\bibinfo{person}{Ana Rebelo}, \bibinfo{person}{Ichiro
  Fujinaga}, \bibinfo{person}{Filipe Paszkiewicz}, \bibinfo{person}{Andre~RS
  Marcal}, \bibinfo{person}{Carlos Guedes}, {and} \bibinfo{person}{Jaime~S
  Cardoso}.} \bibinfo{year}{2012}\natexlab{}.
\newblock \showarticletitle{Optical music recognition: state-of-the-art and
  open issues}.
\newblock \bibinfo{journal}{\emph{International Journal of Multimedia
  Information Retrieval}} \bibinfo{volume}{1}, \bibinfo{number}{3}
  (\bibinfo{year}{2012}), \bibinfo{pages}{173--190}.
\newblock


\bibitem[\protect\citeauthoryear{Shi, Bai, and Yao}{Shi et~al\mbox{.}}{2016}]%
        {shi2016end}
\bibfield{author}{\bibinfo{person}{Baoguang Shi}, \bibinfo{person}{Xiang Bai},
  {and} \bibinfo{person}{Cong Yao}.} \bibinfo{year}{2016}\natexlab{}.
\newblock \showarticletitle{An end-to-end trainable neural network for
  image-based sequence recognition and its application to scene Text
  recognition}.
\newblock \bibinfo{journal}{\emph{IEEE transactions on pattern analysis and
  machine intelligence}} \bibinfo{volume}{39}, \bibinfo{number}{11}
  (\bibinfo{year}{2016}), \bibinfo{pages}{2298--2304}.
\newblock


\bibitem[\protect\citeauthoryear{Su, Lu, Pal, and Tan}{Su
  et~al\mbox{.}}{2012}]%
        {su2012effective}
\bibfield{author}{\bibinfo{person}{Bolan Su}, \bibinfo{person}{Shijian Lu},
  \bibinfo{person}{Umapada Pal}, {and} \bibinfo{person}{Chew~Lim Tan}.}
  \bibinfo{year}{2012}\natexlab{}.
\newblock \showarticletitle{An effective staff detection and removal technique
  for musical documents}. In \bibinfo{booktitle}{\emph{2012 10th IAPR
  International Workshop on Document Analysis Systems}}. IEEE,
  \bibinfo{pages}{160--164}.
\newblock


\bibitem[\protect\citeauthoryear{Vo, Kim, Yang, and Lee}{Vo
  et~al\mbox{.}}{2016}]%
        {vo2016mrf}
\bibfield{author}{\bibinfo{person}{Quang~Nhat Vo}, \bibinfo{person}{Soo~Hyung
  Kim}, \bibinfo{person}{Hyung~Jeong Yang}, {and} \bibinfo{person}{Gueesang
  Lee}.} \bibinfo{year}{2016}\natexlab{}.
\newblock \showarticletitle{An MRF model for binarization of music scores with
  complex background}.
\newblock \bibinfo{journal}{\emph{Pattern Recognition Letters}}
  \bibinfo{volume}{69} (\bibinfo{year}{2016}), \bibinfo{pages}{88--95}.
\newblock


\end{thebibliography}


\end{document}